\documentclass[10pt,journal,compsoc]{IEEEtran}

\ifCLASSOPTIONcompsoc
  \usepackage[nocompress]{cite}
\else
  \usepackage{cite}
\fi

\ifCLASSINFOpdf
\else
\fi

\usepackage{graphicx}
\usepackage[cmex10]{amsmath}
\usepackage{subfig}
\usepackage{array}
\usepackage{booktabs}
\usepackage{multirow}
\usepackage{threeparttable}
\usepackage{amsfonts}
\usepackage{tabularx}
\usepackage{algorithm}
\usepackage{algorithmic}
\usepackage{amsmath}
\usepackage{amssymb}
\DeclareMathOperator*{\argmax}{arg\,max}
\DeclareMathOperator*{\argmin}{arg\,min}
\usepackage{mathtools}
\usepackage{cuted}

\DeclarePairedDelimiter\abs{\lvert}{\rvert}%
\newcolumntype{Y}{>{\centering\arraybackslash}X}
\newcolumntype{b}{>{\centering\arraybackslash}X}
\newcolumntype{s}{>{\hsize=.33\hsize\centering\arraybackslash}X}

\makeatletter
\newcommand*{\rom}[1]{\expandafter\@slowromancap\romannumeral #1@}
\makeatother

\begin{document}
%
\title{Constrained Structure Learning for Scene Graph Generation}
%
%
%
%

\author{Daqi~Liu,~Miroslaw~Bober,~\IEEEmembership{Member,~IEEE},~Josef~Kittler,~\IEEEmembership{Life Member,~IEEE}
\IEEEcompsocitemizethanks{\IEEEcompsocthanksitem The authors are with the Centre for Vision, Speech and Signal Processing, University of Surrey, Guildford GU2 7XH, U.K.  \textbf{Under Review...} \protect\\
E-mail: \{daqi.liu, m.bober, j.kittler\}@surrey.ac.uk}
}

%

\IEEEtitleabstractindextext{%
\begin{abstract}
As a structured prediction task, scene graph generation aims to build a visually-grounded scene graph to explicitly model objects and their relationships in an input image.  Currently, the mean field variational Bayesian framework is the de facto methodology used by the existing methods, in which the unconstrained inference step is often implemented by a message passing neural network. However, such formulation fails to explore other inference strategies, and largely ignores the more general constrained optimization models. In this paper, we present a constrained structure learning method, for which an explicit constrained variational inference objective is proposed. Instead of applying the ubiquitous message-passing strategy, a generic constrained optimization method - entropic mirror descent - is utilized to solve the constrained variational inference step. We validate the proposed generic model on various popular scene graph generation benchmarks and show that it outperforms the state-of-the-art methods.
\end{abstract}

}

\maketitle

\IEEEdisplaynontitleabstractindextext

%
\IEEEpeerreviewmaketitle

\IEEEraisesectionheading{\section{Introduction}\label{sec:introduction}}

%
%
%
%
\IEEEPARstart{S}{cene} graph generation (SGG) task involves building a visually-grounded scene graph to explicitly model objects and their relationships in an input image. Its aim is to facilitate downstream vision tasks such as image captioning \cite{anderson2018bottom}, \cite{yang2019auto} and visual question answering \cite{teney2017graph}, \cite{shi2019explainable}. As a structured prediction task, SGG is generally NP-hard, owing to the exponential complexity of interactions among the output variables (which are expected to form coherent visual relationships) present a huge challenge for directly computing the desired statistics, i.e. the underlying posterior or the relevant marginals. Currently, only pairwise interactions are considered in the SGG task and they are often formulated as triplet structures, in which each triplet consists of three components: a subject, a predicate and an object. 

More specifically, given an input image $x$, a specific type of approximation strategies - variational Bayesian (VB) \cite{wainwright2008graphical}, \cite{fox2012tutorial} - is often applied to accomplish the SGG generation task in the current methods. In this approach, the variational inference step aims to infer the optimum interpretation $z^*$ by means of a max  aposteriori (MAP) estimation strategy, i.e.  $z^*=\argmax_{z}p(z|x)$, while the variational learning step tries to fit the model posterior $p(z|x)$ with the underlying ground-truth posterior $p_r(z|x)$ by maximizing the conditional likelihood. Such a VB framework is implemented in current SGG models \cite{xu2017scene}, \cite{li2017scene}, \cite{dai2017detecting}, \cite{woo2018linknet}, \cite{yang2018graph}, \cite{wang2019exploring}, \cite{tang2019learning}, \cite{li2021bipartite} by constructing two fundamental modules, namely: visual perception and visual context reasoning \cite{liu2019visual}, as shown in Fig.1. For the variational inference step, visual perception initializes the output interpretations, while visual context reasoning refines the interpretations according to certain inference strategies. For the variational learning step, both modules are updated to fit the ground-truth training samples, and the updated modules are applied in the following variational inference step. The resulting optimum interpretation $z^*$ is generated by alternating between the above varitional inference and learning steps.

To construct efficient VB frameworks for complex SGG tasks, the variational distribution $q(z)$ in the current SGG models is often assumed to be fully decomposable. The resulting framework is also known as mean field variational Bayesian (MFVB) \cite{wainwright2008graphical}, \cite{fox2012tutorial}. Such formulation ignores the higher-order interactions in the underlying posterior and it is essentially a locally consistent (rather than globally consistent) approximation of the underlying posterior. However, the MFVB framework is easy to scale to huge datasets without sacrificing much performance, especially when used with stochastic learning methods. This explains the facts why almost all the current SGG methods choose this specific type of VB models as their backbone framework.

In the current MFVB-based SGG models, almost all of them choose a specific optimization strategy - message passing \cite{scarselli2008graph}, \cite{gilmer2017neural}, \cite{wang2018non}, \cite{zhou2020graph} - to infer the optimum interpretations within the variational inference step, and it has became the de facto inference method.
Specifically, different message passing neural network structures \cite{tang2019learning}, \cite{li2021bipartite}, \cite{li2018factorizable}, \cite{chen2019knowledge}, \cite{lin2020gps} have been proposed in recent years to model the above MFVB models and showed to achieve reasonable graph generation performance. However, these methods fail to explore other inference methodologies. More importantly, their variational inference objectives are unconstrained. More generic constrained variational inference objectives have not been properly investigated in the current SGG literature.

To this end, a generic and efficient constrained structure learning (CSL) method is proposed in this paper to solve the SGG task. Unlike the previous methods, a MFVB framework with explicit variational inference and learning objectives is constructed. Moreover, the proposed method considers the variatinal inference step as a constrained optimization problem, rather than an unconstrained one as in previous algorithms. Within the variational inference step, a generic constrained optimization method - entropic mirror decent \cite{beck2003mirror} - is applied to infer the underlying posterior rather than the ubiquitous message passing strategy. The experimental results obtained on the popular Visual Genome and Open Images V6 benchmarks demonstrate the superiority and efficiency of the proposed CSL method. 
\begin{figure}[!t]
\centering
\includegraphics[width=\linewidth]{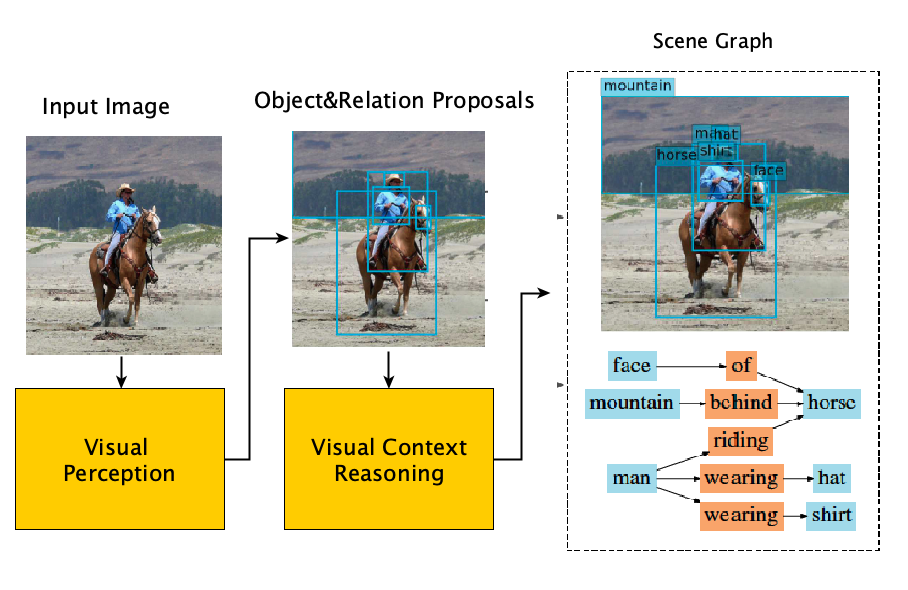}
\caption{ Two fundamental building blocks in SGG: visual perception and visual context reasoning.}
\label{fig_1}
\end{figure} 

This paper is organized as follows: Section 2 presents related works. Section 3 introduces the proposed constrained structure learning methodology. The experimental results and the corresponding analysis are elaborated in Section 4. Finally, the conclusions are drawn in Section 5.

\section{Related Works}

Current SGG models aim to find better feature extraction architectures \cite{xu2017scene}, \cite{li2017scene}, \cite{yang2018graph}, \cite{zhu2018deep}, \cite{li2018factorizable}, \cite{zhang2019vrd},  or address the bias of the relationship prediction process, caused by the long-tail data distribution \cite{zellers2018neural}, \cite{chen2019knowledge}, \cite{gu2019scene}, \cite{tang2019learning}, \cite{tang2020unbiased}, \cite{guo2021general}. Besides \cite{tang2020unbiased}, which utilizes a causal inference, 
almost all of them tend to formulate the SGG task using a mean field variational Bayesian framework. Specifically, the unconstrained variational inference objective is generally minimized by means  of message-passing neural network structures, while the classical cross-entropy loss is often applied to train the associated learning frameworks. Such formulation has became a universal corner stone for almost all the current SGG tasks. In contrast, our proposed method presents an alternative SGG methodology, which constructs a constrained variational inference objective, and applies generic constrained optimization algorithms, rather than message-passing, to infer the optimum interpretation. It has been developed by investigating  generic constrained optimization scenarios and by exploring the alternative inference strategies, which would further improve the applicability and diversity of the SGG methods.

Since the explicit variational inference objective is not required in message-passing based MFVB frameworks, the current SGG models do not need to specify the energy function or the scoring function for the input image $x$ and the output interpretation $z$. More specifically, given the input and output variables, energy function measures their dissimilarities, while scoring function gauges the corresponding similarities. Energy-based models (EBMs) \cite{ranzato2007efficient}, \cite{zhao2017energy} aim to capture the dependencies among variables by associating a scalar energy to each potential configuration of the variables, which is generally non-probabilistic and can be converted to a probabilistic model, assuming the partition function can easily be computed or approximated. Such energy-based formulation is rarely investigated in the current SGG literature and it is only explored by one recently proposed method \cite{suhail2021energy}. However, the contrastive divergence loss applied in the above method may have mode collapse issue \cite{lecun2006tutorial}, which could underestimate the underlying posterior. Unlike the above energy-based algorithm, the proposed method approximates the associated partition function within the proposed MFVB framework. 

From a broad perspective, scene graph generation is a type of structured prediction tasks, which naturally inherits its unique properties and solutions. Traditional techniques like Conditional Random Field (CRF) \cite{sutton2006introduction} or Structured Support Vector Machine (SSVM) \cite{tsochantaridis2005large} provide some basic ways to predict structured outputs $z^*$ from the input image $x$. However, these techniques are quite outdated in the current deep learning era. Therefore, several modern structured prediction methodologies \cite{belanger2016structured}, \cite{graber2018deep}, \cite{graber2019graph} have been proposed in recent years, which leverage the powers of both classical structured prediction techniques and modern deep learning architectures. The representation learning capabilities of these techniques are greatly improved, which paves the way for extending them to more challenging applications. Following this direction, we propose a novel constrained structure learning methodology, which demonstrates its superior scalability and efficiency in complex SGG tasks.
\begin{figure*}[!t]
\centering
\includegraphics[width= 5.5 in]{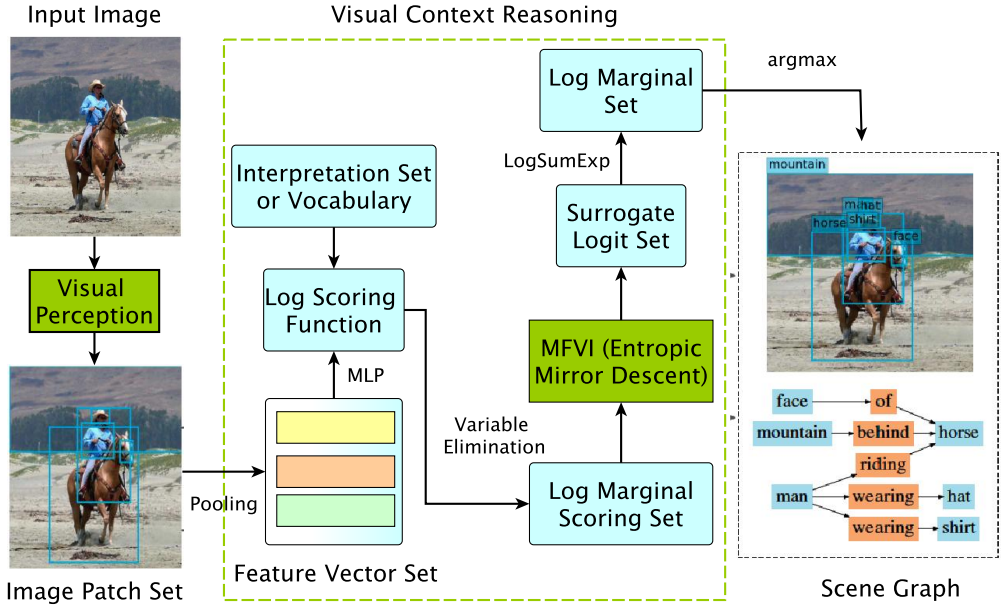}
\caption{ An overview of the proposed method. The green dash line designates the proposed visual context reasoning module. Given an input image, the visual perception module is used to generate a set of region proposals  with the associated image patches. The corresponding feature vectors are obtained by a pooling strategy defined in \cite{ren2015faster}.  Given a feature vector set and candidate interpretations (or vocabulary), the target log scoring function is computed via MLP. The variable elimination technique is applied to infer the corresponding log marginal scoring set, and an MFVI framework with the entropic mirror descent inference method is employed to compute the associated surrogate logit set, which is then transformed into the resulting log marginal set via the $LogSumExp$ trick.  Finally, the output scene graph is generated via the corresponding $\argmax$ operation. A cross-entropy loss is applied in the variational learning step.}
\label{fig_2}
\end{figure*}

\section{Proposed Methodology}

In this section we describe the proposed constrained structure learning method. It is organized as follows: Subsection 3.1 introduces the SGG problem formulation while Subsection 3.2 presents the applied scoring function. The variational Bayesian framework and the specific constrained variational inference strategy are discussed in the last two subsections. A graphical overview of the proposed method is presented in Fig.2.

\subsection{Problem Formulation}

Given an input image $x$, a SGG model aims to build a visually-grounded scene graph by inferring the optimum coherent interpretation $z^*$ for all the  objects and predicates within the input scene. Currently, only pairwise interactions are considered in the output scene graph, which consists of a list of intertwined semantic triplet structures, with each represented as $<s, p, o>$, where $s$ and $o$ are the associated subject and object, while $p$ is the corresponding predicate to describe the relationship between $s$ and $o$. In the current SGG approaches, the supporting evidence for the potential objects are captured by the associated bounding boxes, while their relationships are characterised by the observation in the  corresponding union bounding boxes. The ground-truth training samples are represented as $(\hat{x}_i, \hat{b}_i, \hat{z}_i), i=1,2,...,M$, where $M$ is the number of input images, $\hat{b}_i$ a list of ground-truth bounding boxes for potential objects in image $\hat{x}_i$, and $\hat{z}_i$ is a list  of ground-truth labels for the objects and predicates in image $\hat{x}_i$.

To generate the underlying scene graph, two essential modules are required, namely, visual perception and visual context reasoning modules. The visual perception module aims to locate and instantiate the potential objects and predicates within the input scene, while the visual context reasoning subsystem tries to infer the corresponding interpretations for these objects/predicates using certain inference strategies. In the current SGG tasks, a region proposal network (e.g. faster R-CNN \cite{ren2015faster}) with a VGG-16 \cite{simonyan2014very} or ResNet-101 \cite{he2016deep} backbone is often applied to implement the visual perception module, while the MAP inference is generally deployed to model the visual context reasoning module.

Given an input image $x$, the aim of the visual perception module is to output a set of object region proposals $b^o_i \in \mathbb{R}^4, i=1,2,...,m$, as well as a set of predicate region proposals $b^p_j \in \mathbb{R}^4, j=1,2,...,n$, where $m$ and $n$ are the number of the potential objects and predicates within the input image, respectively. Specifically, suppose $m$ objects are detected in an input image,  a quadratic number of predicate proposals ($m^2-m$) could potentially be generated by computing the  pair of object proposal regions. In reality, the number of predicate proposals $n<<m^2-m$ is much less, and the specific number is purely dependent on the underlying scene graph node adjacency structure of the ground-truth training samples. With the above region proposal sets, the input image $x$ can be divided into two sets of image patches $x_i^o, i=1,2,...,m$ and $x_j^p, j=1,2,...,n$, respectively. Each of these image patches includes all the input pixels defined by its generating region proposals. A pooling strategy (e.g. ROI pooling) is applied to extract the corresponding feature representation sets $y^o_i \in \mathbb{R}^d, i=1,2,...,m$ and $y^p_j \in \mathbb{R}^d, j=1,2,...,n$. 

Given a set of object classes $\mathcal{C}$ and a set of relationship types $\mathcal{R}$, a visual context reasoning module aims to infer a set of object labels (interpretations) $z^o_i \in \mathcal{C} , i=1,2,...,m$ for the input image patch set $x^o_i , i=1,2,...,m$, as well as a set of predicate labels  $z^p_j \in \mathcal{R} , j=1,2,...,n$ for the input predicate image patch set $x^p_j, j=1,2,...,n$. 

With the above traditional SGG formulation, the global contextual information is largely ignored and only local contextual information is considered. In contrast, in our approach, a global latent feature representation set $y_k^g, k=1,2,...,t$, is incorporated into the proposed SGG framework, where $t$ is the number of  input images in a mini-batch. The associated input global image patch set is represented as  $x_k^g, k=1,2,...,t$. The corresponding global region proposals $b_k^g, k=1,2,...,t$ are obtained by finding the unions of all the relevant objects and predicates within the input images. Given a set of global classes $\mathcal{G}$, although the corresponding interpretation set $z_k^g\in \mathcal{G}, k=1,2,...,t$ is not required, 
it is beneficial to incorporate such global contextual information, since it can consider higher-order interactions among the output variables.

\subsection{Scoring Function}
Unlike the previous message-passing based SGG models, an explicit variational inference objective is required in the proposed method. This calls for defining a prior analytical scoring function $s_{\theta}(x, z)$, where $\theta$ denotes the scoring function parameters, and $z$ is the  interpretation of the input image $x$. Such scoring function is generally applied to model the dependencies among the relevant variables. For an undirected graphical model, the scoring function can generally be represented as:
\begin{equation}
s_{\theta}(x,z)=\displaystyle\prod_{r\in R}f_r(x_r, z_r)
\end{equation} 
where $r$ is a clique within a clique list $R$, $f_r$ is a factor function describing the dependencies among the input image patch set $x_r$ and the associated output interpretation $z_r$.  

In the current SGG tasks, only two types of factor functions are considered: the unary factor function $f_u$ and the binary factor function $f_b$. The former gauges the consistency between the input $x$ and the label of a specific node, while the latter characterizes the interactions between a pair of nodes. In this paper, unless indicated otherwise, the discrete label $z$ is generally represented as a corresponding one-hot vector, in which all the elements are set to zeros, except the one corresponding to the correct category. 

To avoid a computationally intractable variational inference objective, $f_r$ is generally formulated as an exponential function and the corresponding log scoring function becomes:
\begin{equation}
logs_{\theta}(x,z)=-\displaystyle\sum_{r\in R}\psi_r(x_r, z_r)
\end{equation}
where $\psi_r$ is the corresponding potential function for the associated clique $r$. Generally, SGG tasks have two types of potential functions: the unary potential function $\psi_u$ and the binary potential function $\psi_b$. The resulting posterior is computed as follows:
\begin{equation}
p_{\theta}(z|x)=\frac{s_{\theta}(x,z)}{s_{\theta}(x)}
\end{equation}
where $s_{\theta}(x)$ is the associated partition function or normalizing constant, and $p_{\theta}(z|x)$ is essentially a Gibbs distribution.

The log scoring function is defined as follows:
\begin{equation}
\begin{split}
logs_{\theta}(x,z)= -\displaystyle\sum_{i=1}^{m}[\psi_u^o(x_i^o, z_i^o)+
\displaystyle\sum_{j\in N(i)}\psi_{b}^o(x_i^o, x_j^p,z_i^o, z_j^p)\\ +\displaystyle\sum_{l\in N(i)}\psi_{b}^o(x_i^o, x_l^o,z_i^o, z_l^o)
+\displaystyle\sum_{k\in N(i)}\psi_{b}^g(x_i^o, x_k^g,z_i^o, z_k^g)]\\
-\displaystyle\sum_{j=1}^{n}[\psi_u^p(x_j^p, z_j^p)+
\displaystyle\sum_{i\in N(j)}\psi_{b}^p(x_i^o, x_j^p,z_i^o, z_j^p)\\
+\displaystyle\sum_{k\in N(j)}\psi_{b}^g(x_j^p, x_k^g,z_j^p, z_k^g)]
\end{split}
\end{equation}
where the superscripts $o$, $p$, $g$ represent the object, the predicate and the global context, respectively. $N(i)$ means the set of neighbouring nodes of the target node $i$, e.g. $l\in N(i)$ is a hypothesised object, linked to the target node i. $m$ and $n$ are the number of objects and predicates detected in an input image, respectively. For objects, such scoring function considers three types of pairwise interactions:  $<object, predicate>$, $<subject, object>$ and $<object, global>$. For predicates, two types of pairwise interactions are considered: $<predicate, object>$ and $<predicate, global>$.

\subsection{Mean Field Variational Bayesian}

The computational complexity of scene graph generation is generally NP-hard, since it is computationally intractable to integrate the exponentially growing number of structured outputs. For this reason, the existing SGG models tend to rely on a specific type of approximation strategies - variational Bayesian \cite{wainwright2008graphical}, \cite{fox2012tutorial} - to estimate the underlying posterior $p_{\theta}(z|x)$ and infer the optimal interpretation $z^*$ for an input image $x$. A  variational Bayesian (VB) model construction includes two alternating steps: variational inference and variational learning, in which the former aims to estimate the underlying posterior $p_{\theta}(z|x)$ with a tractable variational distribution $q(z)$, while the latter tries to fit the underlying posterior with the ground-truth data distribution $p_r(z|x)$, i.e. 
\begin{equation}
\begin{split}
q^*=\argmin_{q}\mathbb{D}(q(z), p_{\theta}(z|x)) \\
\theta^*=\argmin_{\theta}\mathbb{D}(p_r(z|x), p_{\theta}(z|x))
\end{split}
\end{equation} 
where $\mathbb{D}$ is a divergence metric, normally chosen in the form of  KL divergence. The optimum $q^*$ and $\theta ^*$are obtained by alternating between the above two divergence  minimization steps, in which the first step performs variational inference, while the variational learning is executed in the second step.

However, it is impossible to infer the optimum $q^*$ by directly applying the first divergence metric minimization step, as it includes the computationally intractable posterior $p_{\theta}(z|x)$. Luckily, its dual problem - maximizing evidence lower bound (ELBO) - can be readily solved. Based on the Jensen's inequality, the following equation can readily be derived:
\begin{equation}
\begin{split}
logs_{\theta}(x)=\mathbb{E}_{q(z)}log\frac{s_{\theta}(x,z)}{q(z)}+\mathbb{E}_{q(z)}log\frac{q(z)}{p_{\theta}(z|x)}
\end{split}
\end{equation}
where, on the right-hand side, the first term is the so-called ELBO while the second term is the KL divergence between the variational distribution $q(z)$ and the underlying posterior $p_{\theta}(z|x)$. The term on the left-hand side is the log partition function, which is generally computationally intractable. Thus, maximizing ELBO has two consequences: 1) the KL divergence $\mathbb{D}(q(z), p_{\theta}(z|x))$ is minimized; 2) the resulting ELBO becomes a tighter lower bound of $logs_{\theta}(x)$. Therefore, the maximization of ELBO is commonly applied to approximate the computationally intractable log partition function in the variational inference models.

For computational efficiency, the variational distribution $q(z)$ is generally assumed to be fully decomposed in the existing SGG models and each local variational distribution $q_i(z_i)$ is chosen from the conditionally conjugate exponential family \cite{wainwright2008graphical} (categorical distribution for discrete output variables):
\begin{equation}
\begin{split}
q(z)=\displaystyle\prod_{i=1}^{m}q^o_i(z_i^o)\displaystyle\prod_{j=1}^{n}q^p_j(z_j^p)
\end{split}
\end{equation}
where $q^o_i(z_i^o) \in \Delta ^{v_o-1}$ and  $q^p_j(z_j^p) \in \Delta ^{v_p-1}$ ($\Delta$ represents a probability simplex) are local variational distributions for the objects and predicates in the output scene graph, respectively. $v_o$ and $v_p$ are the sizes of vocabularies for the objects and predicates, respectively. With such an assumption, the resulting variational Bayesian model is also known as mean field variational Bayesian (MFVB) \cite{wainwright2008graphical}, \cite{fox2012tutorial} , and the associated inference step is often called mean field variational inference (MFVI). 

In MFVI, the indices of the maximal values of the marginals are exactly the same as the MAP inference results (which is not the case in general). Thus, the target MAP inference in SGG can be transformed into a corresponding marginal inference. In this paper, variable elimination technique \cite{wainwright2008graphical} is applied to infer the associated marginals. Now, for a potential regional proposal $b_i$, delineating the input image patch $x_i$,  the corresponding log marginal distribution is: 
\begin{equation}
\begin{split}
logp_{\theta}(z_i|x_i)&=logs_{\theta}(x_i,z_i)-logs_{\theta}(x_i)\\ 
&\propto \sum_{z\backslash i}[logs_{\theta}(x_i,z)]-logs_{\theta}(x_i)
\end{split}
\end{equation}
where $\sum_{z\backslash i}$ represents marginalization over the interpretations of all the potential output nodes, except the target node $i$, $logs_{\theta}(x_i, z_i)$ is the associated log marginal scoring function, and $logs_{\theta}(x_i)$ stands for the partition function associated with $x_i$. 

Specifically, given a potential object/predicate regional proposal $b^o_i$/$b_j^p$, the corresponding log marginal scoring function is computed as follows:
\begin{equation}
\begin{split}
logs_{\theta}(x_i^o,z_i^o)\propto -[\psi_u^o(x_i^o, z_i^o)+\sum_{j\in N(i)} m^{op}_{j\to i}\\
+\sum_{l\in N(i)} m^{oo}_{l\to i}+\sum_{k\in N(i)} m^{og}_{k\to i}] \\
logs_{\theta}(x_j^p,z_j^p)\propto -[\psi_u^p(x_j^p, z_j^p)+\sum_{i\in N(j)} m^{po}_{i\to j}\\
+\sum_{k\in N(j)} m^{pg}_{k\to j}] \\
m^{op}_{j\to i}=\sum_{z_j^p\in \mathcal{R}}\psi^o_b(x_i^o, x_j^p, z_i^o, z_j^p)\\
m^{oo}_{l\to i}=\sum_{z_l^o\in \mathcal{C}}\psi^o_b(x_i^o, x_l^o, z_i^o, z_l^o)\\
m^{og}_{k\to i}=\sum_{z_k^g\in \mathcal{G}}\psi^o_b(x_i^o, x_k^g, z_i^o, z_k^g)\\
m^{po}_{i\to j}=\sum_{z_i^o\in \mathcal{C}}\psi^p_b(x_i^o, x_j^p, z_i^o, z_j^p)\\
m^{pg}_{k\to j}=\sum_{z_k^g\in \mathcal{G}}\psi^p_b(x_k^g, x_j^p, z_k^g, z_j^p)
\end{split}
\end{equation}
where the above associated functions are defined as follows:
\begin{equation}
\begin{split}
\label{ref1}
\psi_u^o(x_i, z_i^o) &= h^o_{\theta}(x_i)\cdot z_i^o\\
\psi_u^p(x_j, z_j^p) &= h^p_{\theta}(x_j)\cdot z_j^p\\
m^{op}_{j\to i}(x_i^o, x_j^p, z_i^o, z_j^p) &= g^{op}_{\theta}(x_i^o, x_j^p)\cdot z_i^o\\
m^{oo}_{l\to i}(x_i^o, x_l^o, z_i^o, z_l^o) &= g^{oo}_{\theta}(x_i^o, x_l^o)\cdot z_i^o\\
m^{og}_{k\to i}(x_i^o, x_k^g, z_i^o, z_k^g) &= g^{og}_{\theta}(x_i^o, x_k^g)\cdot z_i^o\\
m^{po}_{i\to j}(x_i^o, x_j^p,  z_i^o,  z_j^p) &= g^{po}_{\theta}(x_i^o, x_j^p)\cdot z_j^p\\
m^{pg}_{k\to j}(x_j^p, x_k^g,  z_j^p,  z_k^g) &= g^{pg}_{\theta}(x_j^p, x_k^g)\cdot z_j^p
\end{split}
\end{equation}
In \eqref{ref1}, $\cdot$ means an inner product, $z_i^o$/$z_j^p$ is a one-hot representation of a potential object/predicate in an input image $x$. The feature representation learning functions $h^o_{\theta}$, $h^p_{\theta}$, $g^{op}_{\theta}$, $g^{oo}_{\theta}$, $g^{og}_{\theta}$, $g^{po}_{\theta}$, $g^{pg}_{\theta}$ are constructed by combing visual perception modules and multi-layer perceptrons (MLPs), which are parameterized by $\theta$. As indicated in Fig.3, each of these functions will first map the input image patches $x$ into the corresponding feature representations $y\in \mathbb{R}^d$ via the visual perception module, and then obtain the resulting $\mathbb{R}^v$ dimensional feature vector by feeding relevant $y$ into the MLP. The output log score is the inner product of the above $\mathbb{R}^v$ dimensional feature vector and the corresponding $v$-dimensional one-hot vector $z$.

To infer the target log marginal $logp_{\theta}(z_i|x_i)$, besides the above $logs_{\theta}(x_i, z_i)$, it is necessary to estimate the computationally intractable $logs_{\theta}(x_i)$. To this end, an explicit constrained variational inference objective is proposed:
\begin{equation}
\begin{split}
logs_{\theta}(x_i)\triangleq \max_{q_i}\mathbb{L}(q_i)&=\max_{q_i}\mathbb{E}_{q_i(z_i)}log\frac{s_{\theta}(x_i,z_i)}{q_i(z_i)}\\
& s.t. \; \; q_i(z_i) \in \Delta^{v-1}
\end{split}
\end{equation}
where $\Delta^{v-1}$ is a $v-1$ simplex and $\mathbb{L}(q_i)$ represents the variational inference objective. Unlike the previous SGG models, the variational inference step in the proposed method becomes explicit and is formulated as a constrained maximization problem. Specifically, for a potential object/predicate regional proposal $b^o_i$/$b_j^p$ , its associated variational inference objective is as follows:
\begin{equation}
\begin{split}
\mathbb{L}^o(q_i)\propto -\mathbb{E}_{q_i(z_i^o)}[\psi_u^o(x_i^o, z_i^o)+\sum_{j\in N(i)} m^{op}_{j\to i} \\
+\sum_{l\in N(i)} m^{oo}_{l\to i}+\sum_{k\in N(i)} m^{og}_{k\to i}]- \mathbb{E}_{q_i(z_i^o)}logq_i(z_i^o)\\
 s.t. \; \; q_i(z_i^o) \in \Delta^{v_o-1} \\
\mathbb{L}^p(q_j)\propto -\mathbb{E}_{q_j(z_j^p)}[\psi_u^p(x_j^p, z_j^p)+\sum_{i\in N(j)} m^{po}_{i\to j} \\
+\sum_{k\in N(j)} m^{pg}_{k\to j}] - \mathbb{E}_{q_j(z_j^p)}logq_j(z_j^p)\\
 s.t. \; \; q_j(z_j^p) \in \Delta^{v_p-1}
\end{split}
\end{equation}
where $\mathbb{L}^o$/$\mathbb{L}^p$represents the object/predicate variational inference objective, and $v_o$/$v_p$ is the object/predicate vocabulary size. Furthermore, the target log probability (or logit) $logp_{\theta}(z_i|x_i)$ is computed via a surrogate logit $\phi$:
\begin{equation}
\begin{split}
&logp_{\theta}(z_i|x_i)\triangleq \phi+C\\
\phi= [logs_{\theta}(x_i,z_i) &-\max_{q_i}\mathbb{E}_{q_i(z_i)}log\frac{s_{\theta}(x_i,z_i)}{q_i(z_i)}]
\end{split}
\end{equation}
where $C$ is an associated constant w.r.t. $x_i$ and $z_i$. Using the $LogSumExp$ trick, we can compute $logp_{\theta}(z_i|x_i)$ by omitting the above constant $C$: 
\begin{equation}
\begin{split}
logp_{\theta}(z_i|x_i)\triangleq \phi - log{\lVert e^{\phi} \rVert_1}
\end{split}
\end{equation}
where, for an input image patch $x_i$, its optimum interpretation is computed as $z_i^*=\argmax_{z_i}logp_{\theta}(z_i|x_i)$. For discrete output variables, $z_i^*$ is the indice of the max value of the log probability $logp_{\theta}(z_i|x_i)$.
\begin{figure}[!t]
\centering
\includegraphics[width=\linewidth]{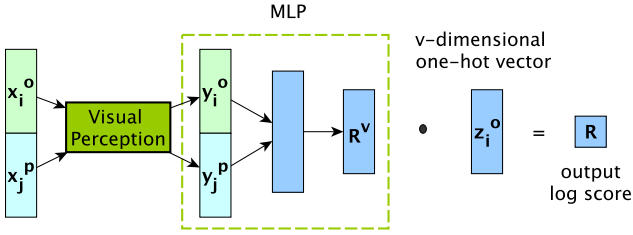}
\caption{ An example three-layer MLP to compute $m^{op}_{j\to i}$. Given image patches $x_i^o$ and $x_j^p$, the visual perception module  computes the corresponding feature representations $y_i^o\in \mathbb{R}^d$ and $y_j^p \in \mathbb{R}^d$, which are then fed into the MLP to obtain a resulting $\mathbb{R}^v$ dimensional feature vector. The output log score is essentially the inner product of the above $\mathbb{R}^v$ dimensional feature vector and the corresponding $v$-dimensional one-hot vector $z_i^o$.}
\label{fig_3}
\end{figure} 

To complete the proposed MFVB framework, cross-entropy loss is applied to implement the associated variational learning step. Specifically, suppose the associated training samples are drawn from a ground-truth data distribution $p_r(z|x)$. Then one can find the optimum parameters $\theta^*$ via:
\begin{equation}
\begin{split}
\theta^*= \argmin_{\theta}\mathbb{L}(\theta) = \argmin_{\theta}\sum_{i=1}^{m+n}\mathbb{E}_{p_r(z_i|x_i)}[-logp_{\theta}(z_i|x_i)]
\end{split}
\end{equation}
where $\mathbb{L}(\theta)$ represents the variational learning objective. 

\subsection{Entropic Mirror Descent Inference Method}

Unlike the previous message-passing based SGG models, the variational inference step in the above proposed MFVB framework is a constrained optimization problem, as demonstrated in Equation (11). Specifically, the variational inference step aims to maximize the associated ELBO $\mathbb{L}(q_i)$, subject to the constraint that the applied variational distribution $q_i(z_i)$ resides in a $v-1$ simplex. 

The projected gradient descent (PGD) methods \cite{eicke1992iteration} are often applied to solve the above constrained optimization problem. Compared with the traditional gradient descent method, it essentially adds a $\mathbb{L}_2$ regularization term in the weight update step, which projects the updated weight to a valid set defined by the constraints. Mirror descent (MD) \cite{nemirovskij1983problem}, \cite{krichene2015accelerated} is a generalized projected gradient descent method in the sense that it replaces the above $\mathbb{L}_2$ Euclidean distance with a more general Bregman distance \cite{teboulle1992entropic}. Since the constraint in the above maximization problem is a probability simplex, the negative entropy $\mathbb{E}_{q_i(z_i)}logq_i(z_i)$ can be used as a specific function to construct the associated Bregman distance. The resulting algorithm is also known as the entropic mirror descent (EMD) \cite{beck2003mirror}.
\begin{algorithm}[ht]
 \caption{Entropic Mirror Descent Inference Method}\label{euclid}
 \textbf{Input} variational distribution $q_i$, number of iterations $T$, an initial learning rate $\alpha$, a predefined objective $\mathbb{L}_p(q_i)$, a small positive value $\epsilon$\\
 \textbf{Output} optimum $q_i^*$
 \begin{algorithmic}[1]
 \STATE randomly initialize $q_i$
 \FOR{iteration $i=1$ to $T$}
 \STATE compute $\mathbb{L}(q_i)$ and its derivative $\bigtriangledown_{q_i}\mathbb{L}(q_i)$
 \STATE set learning rate $\alpha = \alpha/\sqrt{i}$
 \STATE end the loop if {$\abs{\mathbb{L}(q_i)-\mathbb{L}_p(q_i)}<\epsilon$}
 \STATE set $\mathbb{L}_p(q_i)=\mathbb{L}(q_i)$
 \STATE compute $r=\alpha\cdot\bigtriangledown_{q_i}\mathbb{L}(q_i)$
 \STATE compute $r=q_i\cdot e^{r-\max(r)}$
 \STATE set $q_i=\frac{r}{\lVert r \rVert_1}$
 \ENDFOR
 \end{algorithmic}
 \end{algorithm}

The above generic entropic mirror descent method is applied to solve the associated constrained optimization problem formulated in the proposed variational inference step. The proposed entropic mirror descent inference method is summarised in Algorithm 1. Compared with the projected gradient descent algorithms, such method generally converges faster due to the utilization of the geometry of the optimization problem \cite{raskutti2015information}, which is especially desirable in complex SGG tasks. 
 
 \section{Experiments}

To validate the proposed method, in this section, it is  compared with various state-of-the-art models on two popular scene graph generation benchmarks: Visual Genome \cite{krishna2017visual} and Open Images V6 \cite{alina2020open}, respectively. An experimental analysis and ablation study are also presented. Finally, visualization results are provided and discussed in the last subsection. 
\begin{figure}[!t]
\centering
\includegraphics[width=\linewidth]{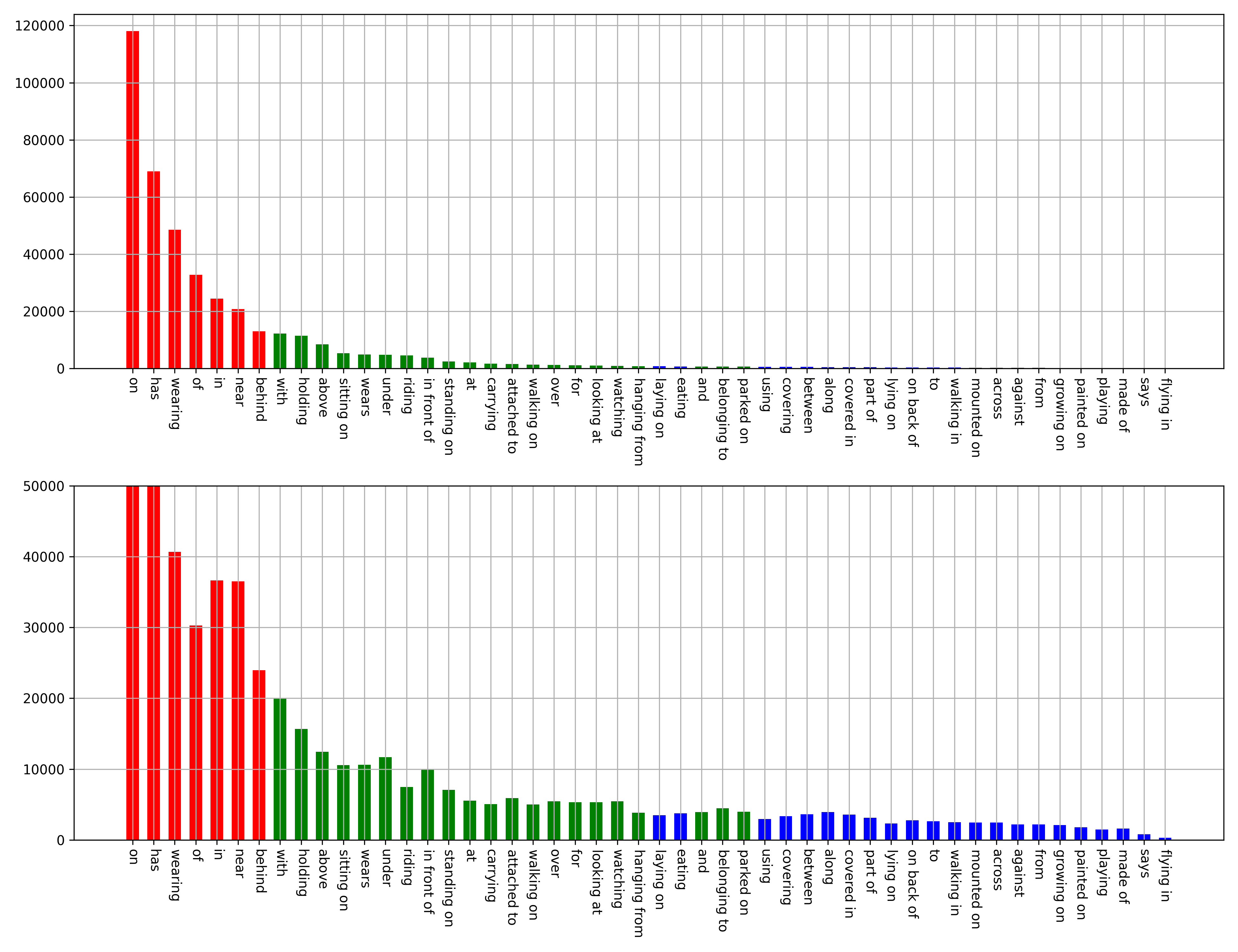}
\caption{ Long-tail category groups in the Visual Genome training split, in which $y$ axis represents the number of samples. In the Visual Genome long-tail data distribution, the predicate categories are divided into three disjoint sets: $head$ (red bars), $body$ (green bars) and $tail$ (blue bars).} 
\label{fig_4}
\end{figure} 

\subsection{Visual Genome}

\subsubsection{Benchmark}

Visual Genome \cite{krishna2017visual} is a predominant SGG benchmark, which contains 108,077 images with an average of 38 objects and 22 relationships per image. We adopt the same data split protocol as \cite{xu2017scene},  in which the most frequent 150 object classes and 50 predicate classes are chosen for the experiment. Specifically, Visual Genome is divided into a training set ($70\%$) and a test set ($30\%$). An evaluation set ($5k$), used for validation, is a random subset of the training set. Moreover, as in \cite{liu2019large}, based on the number of instances in training split, the categories are divided into three disjoint sets: $head$ (more than $10k$), $body$ ($0.5k \sim 10k$) and $tail$ (less than $0.5k$), as demonstrated in Fig.4. 

\subsubsection{Evaluation Metrics}

In this paper, as the evaluation metric we choose the mean Recall$@K$ ($mR@K$) rather than the regular Recall$@K$($R@K$), due to the data imbalance that leads to a bias, as demonstrated in \cite{tang2020unbiased}. In particular, $R@K$ only focuses on common predicates (e.g. $on$), with abundant training samples, and underestimates the informative predicate categories (e.g. $standing\; on$ or $parked\; on$) represented by a fewer  training samples. Like the previous algorithms, we validate the proposed method on the following three settings: 1) Predicate Classification (PredCls), which predicts the predicate labels, given the input image, the ground-truth bounding boxes and object labels; 2) Scene Graph Classification (SGCls), which predicts the labels for objects and predicates, given the input image and the ground-truth bounding boxes; 3) Scene Graph Detection (SGDet), which predicts the scene graph from the input image. 

\subsubsection{Implementation Details}

As in \cite{tang2020unbiased}, in this paper,  ResNeXt-101-FPN \cite{he2016deep} and Faster- RCNN \cite{ren2015faster} are chosen as the backbone and the object detector, respectively for the visual perception module. We choose the step training strategy, in which the pre-trained optimum parameters are loaded into the above models and kept frozen during training. To achieve an effective trade-off between the head and the tail categories, we adopt the same bi-level data resampling strategy as in \cite{li2021bipartite}, which includes image-level over-sampling (the data sampler creates a random permutation of images in which each image is repeated according to its repeat factor $t$ in each epoch) and instance-level under-sampling (the data sampler under-samples based on a drop-out probability for instances of different predicate classes in each image). We set the repeat factor $t=0.07$ and the instance drop rate $\gamma_{d}=0.7$ in this experiment. The batch size $bs$ is set to 12. For the PredCls and SGCls settings, we apply a two-layer MLP to construct the associated log scoring function and use a higher learning rate ($0.008\times bs$) in the SGD optimizer. For the SGDet setting, we employ a three-layer MLP to build the corresponding log scoring function, and utilize a lower learning rate ($0.005\times bs$) in the SGD optimizer. 
\begin{table}[!t]
   \resizebox{\linewidth}{!}{
   \begin{threeparttable}
	\renewcommand{\arraystretch}{1.5}
	\caption{Performance comparison on Visual Genome dataset.}
	\centering
    \begin{tabular}{@{\extracolsep{4pt}}*7c@{}}
	\toprule
	{} & \multicolumn{2}{c}{PredCls} & \multicolumn{2}{c}{SGCls} & \multicolumn{2}{c}{SGDet}\\ \cmidrule{2-3} \cmidrule{4-5} \cmidrule{6-7}
	{Method} & {mR@50} & {mR@100} & {mR@50} & {mR@100} & {mR@50} & {mR@100}\\
	\midrule
    RelDN$^{\dagger}$\cite{zhang2019vrd}  & $15.8$ & $17.2$ & $9.3$ & $9.6$ & $6.0$ & $7.3$\\
	Motifs\cite{zellers2018neural}  & $14.6$ & $15.8$ & $8.0$ & $8.5$ & $5.5$ & $6.8$\\
	Motifs*\cite{zellers2018neural}  & $18.5$ & $20.0$ & $11.1$ & $11.8$ & $8.2$ & $9.7$\\
	G-RCNN$^{\dagger}$\cite{yang2018graph}   & $16.4$ & $17.2$ & $9.0$ & $9.5$ & $5.8$ & $6.6$\\
	MSDN$^{\dagger}$\cite{li2017scene}   & $15.9$ & $17.5$ & $9.3$ & $9.7$ & $6.1$ & $7.2$\\
    VCTree\cite{tang2019learning} & $15.4$ & $16.6$ & $7.4$ & $7.9$ & $6.6$ & $7.7$\\
    GPS-Net$^{\dagger}$\cite{lin2020gps}   & $15.2$ & $16.6$ & $8.5$ & $9.1$ & $6.7$ & $8.6$\\
    GPS-Net$^{\dagger *}$\cite{lin2020gps}   & $19.2$ & $21.4$ & $11.7$ & $12.5$ & $7.4$ & $9.5$\\
    Transformer\cite{vaswani2017attention}  & $16.3$ & $17.6$ & $10.1$ & $10.7$ & $8.1$ & $9.6$\\
    VCTree-TDE\cite{tang2020unbiased}  & $25.4$ & $28.7$ & $12.2$ & $14.0$ & $9.3$ & $11.1$\\
    BGNN\cite{li2021bipartite}  & $30.4$ & $32.9$ & $14.3$ & $16.5$ & $10.7$ & $12.6$\\
    \textbf{CSL} & $\mathbf{29.5}$ & $\mathbf{31.6}$ & $\mathbf{16.7}$ & $\mathbf{17.9}$ & $\mathbf{11.9}$ & $\mathbf{14.3}$\\
	\bottomrule
    \end{tabular}
    \begin{tablenotes}
	\item [\textbullet] Note: All the above methods apply ResNeXt-101-FPN as the backbone. $*$ means the re-sampling strategy \cite{gupta2019lvis} is applied in this method, and $\dagger$ depicts the results reproduced with the latest code from the authors. 
      \end{tablenotes}
    \end{threeparttable}
    }
\end{table} 

\subsubsection{Comparisons with State-of-the-Art Methods}

As demonstrated in Table 1, the proposed CSL method achieves state-of-the-art performance in the SGCls and SGDet settings and comparable performance with the latest BGNN model \cite{li2021bipartite} in the PredCls setting. Specifically, compared with the latest BGNN method, the SGDet performance gain achieved by the proposed method is $11.2\%$ and $13.5\%$, respectively. It is worth noting that the proposed CSL method can achieve such performance with a relatively small number of training iterations, since the generic entropic mirror descent method applied in MFVI converges faster than the message passing strategy. 
\begin{table}[!t]
   \resizebox{\columnwidth}{!}{
   \begin{threeparttable}
	\renewcommand{\arraystretch}{1.5}
	\caption{The performance comparison on Visual Genome for the long-tail category groups in the SGDet setting ($R@100$).}
	\centering
    \begin{tabularx}{\linewidth}{bssss} 
	\toprule
	{Method} & {Head} & {Body} & {Tail} & {Mean}\\ 
	\midrule
	RelDN$^{\dagger}$\cite{zhang2019vrd}  & $34.1$ & $6.6$ & $1.1$ & $13.9$\\
	Motifs\cite{zellers2018neural} & $36.1$ & $7.0$ & $0.0$ & $14.4$ \\
	Motifs*\cite{zellers2018neural} & $34.2$ & $8.6$ & $2.1$ & $15.0$ \\
	G-RCNN$^{\dagger}$\cite{yang2018graph} & $28.6$ & $6.5$ & $0.1$ & $11.7$ \\
	MSDN$^{\dagger}$\cite{li2017scene} & $35.1$ & $5.5$ & $0.0$ & $13.5$ \\
    VCTree-TDE\cite{tang2020unbiased} & $24.5$ & $13.9$ & $0.1$ & $12.8$ \\
    GPS-Net$^{\dagger}$\cite{lin2020gps} & $34.5$ & $7.0$ & $1.0$ & $14.2$ \\
    GPS-Net$^{\dagger *}$\cite{lin2020gps} & $30.4$ & $8.5$ & $3.8$ & $14.2$ \\
    BGNN\cite{li2021bipartite} & $33.4$ & $13.4$ & $6.4$ & $17.7$ \\
    \textbf{CSL} & $\mathbf{33.6}$ & $\mathbf{13.5}$ & $\mathbf{8.8}$ & $\mathbf{18.6}$ \\
	\bottomrule
    \end{tabularx}
    \begin{tablenotes}
	\item [\textbullet] Note: All the above methods apply ResNeXt-101-FPN as the backbone. $*$ means a re-sampling strategy \cite{gupta2019lvis} is applied in this method, and $\dagger$ depicts the results reproduced with the latest code from the authors. 
      \end{tablenotes}
    \end{threeparttable}
    }
\end{table} 

\begin{figure}[!t]
\centering
\includegraphics[width=\linewidth]{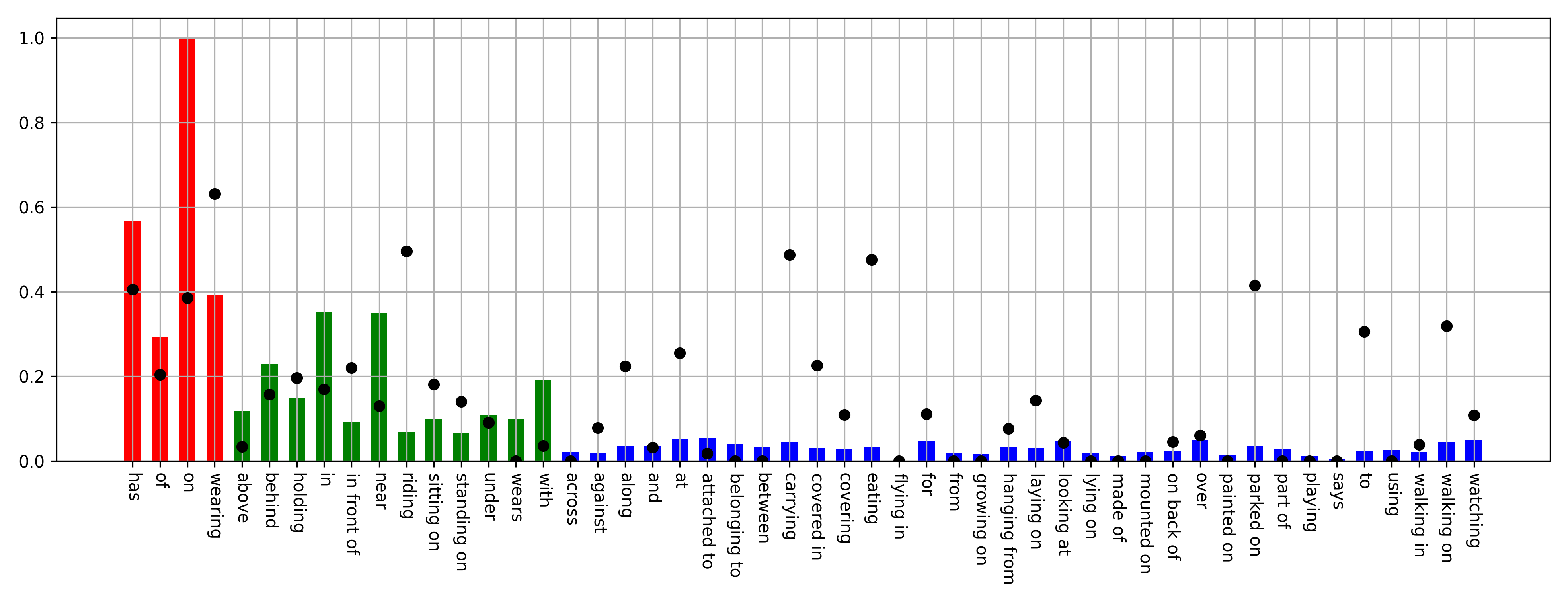}
\caption{ The mean Recall$@100$ performance (denoted by black dot) for each predicate category achieved by our proposed CSL method. The $y$ axis represents the min-max normalized frequency. For the $tail$ categories (blue bars) with a fewer training samples, the proposed CSL method still achieves a reasonable detection rate, which demonstrates its ability to  rectify the biased relationship prediction problem caused by long-tail data distribution. }
\label{fig_5}
\end{figure} 

Moreover, in Table 2, where we compare the performances on long-tail category groups in SGDet setting, the proposed CSL method achieves the best mean performance. More importantly, CSL outperforms the previous methods by a large margin on the $tail$ group, which clearly demonstrate its superior detection capability for the informative predicate categories with a fewer training samples. In other words, unlike the previous models, which mainly detect the dominant predicate categories, the proposed CSL method has the capacity to detect more informative predicate categories and thus reduce the problem of bias in the relationship prediction caused by the long-tail data distribution, as demonstrated in Fig.5. 
\begin{table}[!t]
   \resizebox{\columnwidth}{!}{
   \begin{threeparttable}
	\renewcommand{\arraystretch}{1.5}
	\caption{The performance comparison on the Visual Genome dataset using the balance adjustment strategy.}
	\centering
    \begin{tabular}{@{\extracolsep{4pt}}*7c@{}}
	\toprule
	{} & \multicolumn{2}{c}{PredCls} & \multicolumn{2}{c}{SGCls} & \multicolumn{2}{c}{SGDet}\\ \cmidrule{2-3} \cmidrule{4-5} \cmidrule{6-7}
	{Method} & {mR@50} & {mR@100} & {mR@50} & {mR@100} & {mR@50} & {mR@100}\\
	\midrule
	Motifs+BA\cite{guo2021general}  & $29.7$ & $31.7$ & $16.5$ & $17.5$ & $13.5$ & $15.6$\\
	VCTree+BA\cite{guo2021general}  & $30.6$ & $32.6$ & $20.1$ & $21.2$ & $13.5$ & $15.7$\\
    Transformer+BA\cite{guo2021general}   & $31.9$ & $34.2$ & $18.5$ & $19.4$ & $14.8$ & $17.1$\\
    \textbf{CSL+BA} & $\mathbf{36.9}$ & $\mathbf{39.2}$ & $\mathbf{19.7}$ & $\mathbf{21.2}$ & $\mathbf{15.7}$ & $\mathbf{18.4}$\\
	\bottomrule
    \end{tabular}
    \begin{tablenotes}
	\item [\textbullet] Note: All the above methods apply the same balance adjustment strategy as in \cite{guo2021general} .
      \end{tablenotes}
    \end{threeparttable}
    }
\end{table} 

To improve the performance further, we adopt the generic balance adjustment strategy \cite{guo2021general} into our proposed CSL method and compare the resulting performance with several state-of-the-art models in Table 3.  For a fair comparison, we choose the three baseline models presented in \cite{guo2021general}. The balance adjustment strategy includes two important processes: semantic adjustment and balanced predicate learning. The former aims to cast the common predictions generated by an SGG model as informative ones, while the latter tries to extend the sampling space for the informative predicates. These processes are applied to solve two sub-problems: semantic space imbalance and training sample imbalance.
\begin{figure}[!t]
\centering
\includegraphics[width=\linewidth]{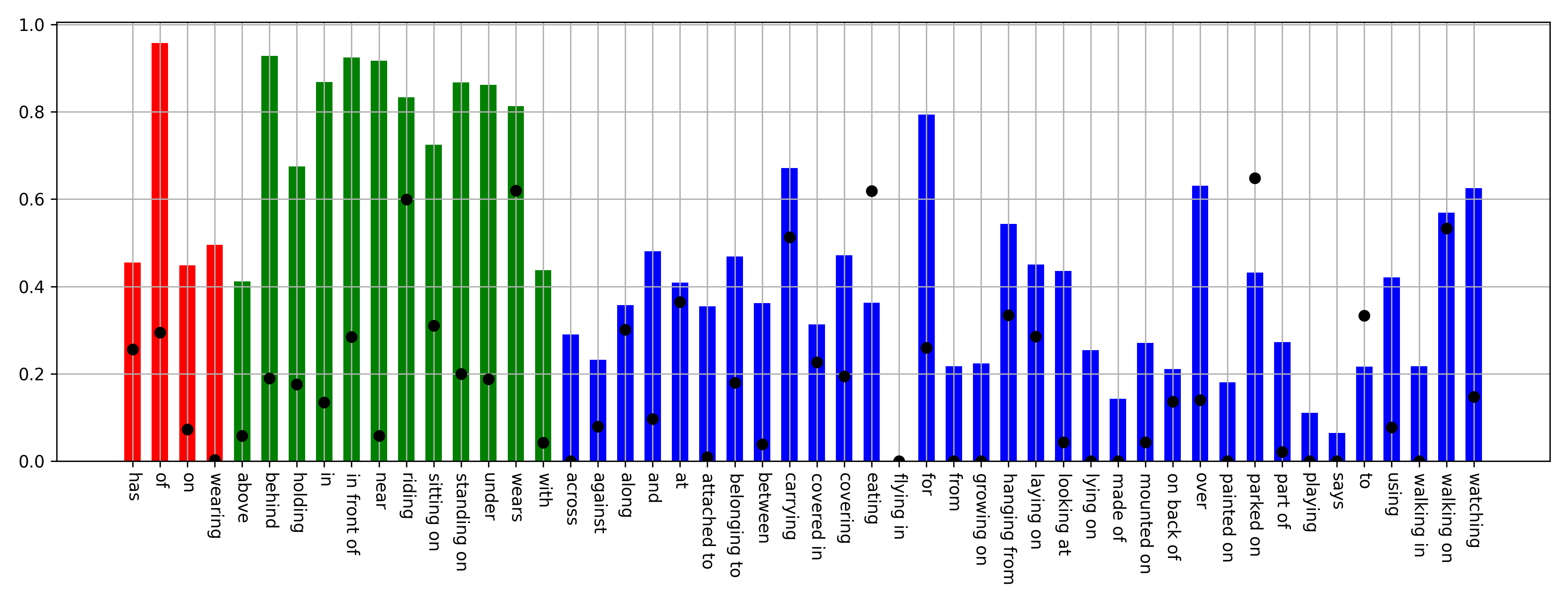}
\caption{ The mean Recall$@100$ performance (denoted as black dot) for each predicate category obtained with our proposed CSL+BA method, in which $y$ axis represents the min-max normalized frequency. Based on the Shannon information theory, the informative $tail$ (blue bars) and $body$ (green bars) predicate categories are largely kept, while the number of samples of the common $head$ (red bars) predicate categories are strictly controlled. Compared with Fig.5, the training samples within the CSL+BA method are more balanced.}
\label{fig_6}
\end{figure} 

As shown in Table 3, the resulting CSL+BA method achieves the state-of-the-art performance on the Visual Genome benchmark. It outperforms the previous models by a large margin, especially for the PredCls setting. As demonstrated in Fig.6, due to the balanced predicate learning, the resulting CSL+BA method has more balanced training samples, in which the more informative (from Shannon information theory perspective) $tail$ and $body$ predicate categories are largely kept, while the common $head$ predicate categories are strictly constrained by means of training sample pruning. With the transition matrix introduced in the semantic adjustment process, the resulting CSL+BA method tends to choose the informative predicates rather than the common ones.  Accordingly, the black dots (representing the mean Recall$@100$) in the $body$ and $tail$ sets of Fig.6 are generally higher than the ones in Fig.5.  

\subsubsection{Ablation Study}

In this section, we investigate the detection performance dependency of the proposed CSL method on the number of iterations $T$ of the entropic mirror descent optimisation procedure, 
and present the results in Table 4. Note, the associated positive value $\epsilon$ of EMD is set to $0.0001$ in this experiment, which is applied for early stopping. Generally, the detection accuracy gradually improves with the number of iterations until convergence. Moreover, the ablation study also reflects the convergence rate of the applied entropic mirror descent method. As shown in Table 4, the applied EMD method exhibits reasonably high convergence rate, requiring only around $10$ iterations to converge. For complex SGG tasks, such high convergence rate is very welcome.
\begin{table}[!t]
   \begin{threeparttable}
	\renewcommand{\arraystretch}{1.5}
	\caption{Ablation study of the impact of the entropic mirror descent method.}
	\centering
    \begin{tabularx}{\linewidth}{bsss}
	\toprule
	{Iteration Numbers $T$} &  {mR@20} & {mR@50} & {mR@100}\\ 
	\midrule
    $5$ & $8.96$ & $11.93$ & $13.92$ \\
	$10$ &  $9.12$ & $12.24$ & $14.44$ \\
    $15$ & $9.23$ & $12.10$ & $14.10$\\
	$20$ & $9.14$ & $12.09$ & $14.34$ \\
	\bottomrule
    \end{tabularx}
    \begin{tablenotes}
	\item [\textbullet] Note: We test the performance of the proposed CSL method using EMD obtained with an increasing number of interactions $T$. 
	The associated small positive value $\epsilon$ in EMD is set to $0.0001$.
      \end{tablenotes}
    \end{threeparttable}
\end{table} 

\subsection{Open Images V6}

\subsubsection{Benchmark}

Open Images V6 \cite{alina2020open} (301 object categories and 31 predicate categories) from Google is another popular SGG benchmark, with a superior annotation quality. The dataset contains 126,368 training images, 1813 validation images and 5322 test images. In this experiment, we choose the same data processing protocols as in \cite{alina2020open}, \cite{zhang2019vrd}, \cite{lin2020gps}.
\begin{table}[!t]
   \resizebox{\columnwidth}{!}{
   \begin{threeparttable}
	\renewcommand{\arraystretch}{1.5}
	\caption{A performance comparison on the Open Images V6 dataset.}
	\centering
    \begin{tabular}{@{\extracolsep{4pt}}*6c@{}}
	\toprule
	{Method} & {mR@50} & {R@50} & {wmAP\_rel} & {wmAP\_phr} & {score\_wtd} \\
	\midrule
	RelDN$^{\dagger}$\cite{zhang2019vrd}  & $33.98$ & $73.08$ & $32.16$ & $33.39$ & $40.84$\\
	RelDN$^{\dagger*}$\cite{zhang2019vrd}  & $37.20$ & $75.34$ & $33.21$ & $34.31$ & $41.97$ \\
    VCTree$^{\dagger}$\cite{tang2019learning} & $33.91$ & $74.08$ & $34.16$ & $33.11$ & $40.21$ \\
    G-RCNN$^{\dagger}$\cite{yang2018graph}   & $34.04$ & $74.51$ & $33.15$ & $34.21$ & $41.84$ \\
	Motifs$^{\dagger}$\cite{zellers2018neural}  & $32.68$ & $71.63$ & $29.91$ & $31.59$ & $38.93$ \\
    VCTree-TDE$^{\dagger}$\cite{tang2020unbiased}  & $35.47$ & $69.30$ & $30.74$ & $32.80$ & $39.27$ \\
    GPS-Net$^{\dagger}$\cite{lin2020gps}   & $35.26$ & $74.81$ & $32.85$ & $33.98$ & $41.69$ \\
    GPS-Net$^{\dagger *}$\cite{lin2020gps}   & $38.93$ & $74.74$ & $32.77$ & $33.87$ & $41.60$ \\
    BGNN\cite{li2021bipartite}  & $40.45$ & $74.98$ & $33.51$ & $34.15$ & $42.06$ \\
    \textbf{CSL} & $\mathbf{41.72}$ & $\mathbf{75.44}$ & $\mathbf{34.30}$ & $\mathbf{35.38}$ & $\mathbf{42.86}$ \\
	\bottomrule
    \end{tabular}
    \begin{tablenotes}
	\item [\textbullet] Note: All the above methods use ResNeXt-101-FPN as the backbone. $*$ means the re-sampling strategy \cite{gupta2019lvis} is applied in this method, and $\dagger$ depicts the  results reproduced using the latest code from the authors. 
      \end{tablenotes}
    \end{threeparttable}
    }
\end{table} 

\begin{figure*}[!t]
\centering
\includegraphics[width= \linewidth]{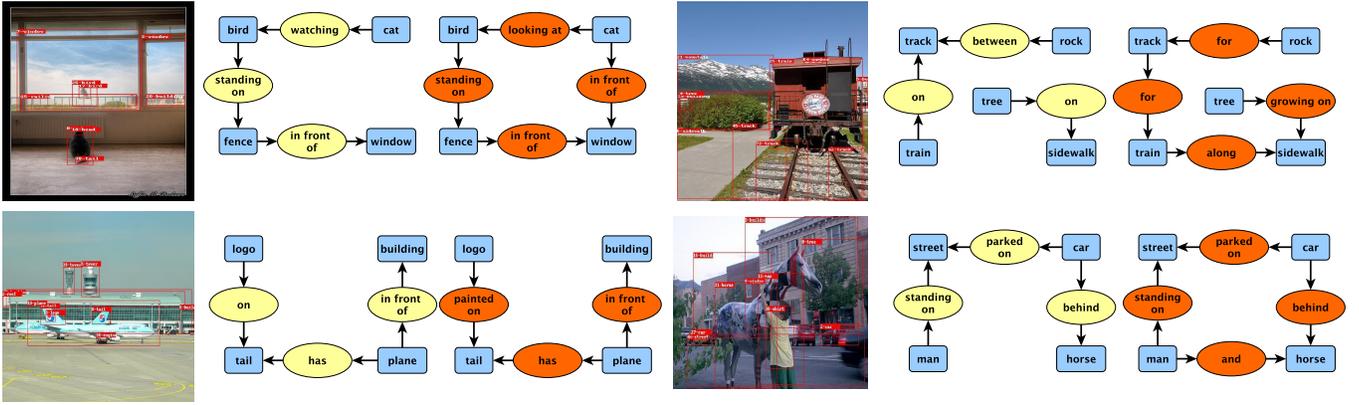}
\caption{ Visualization of the results produced by the  proposed CSL method (in yellow) as well as the corresponding CSL+BA algorithm (in orange). Due to the limited space, only the top predicates are shown in this image.  Unlike the traditional SGG models, the proposed CSL method is able to detect informative predicate categories (i.e. $standing\; on$ rather than $on$), and  the corresponding CSL+BA algorithm further improves ability with the help of the balance adjustment strategy.}
\label{fig_7}
\end{figure*} 

\subsubsection{Evaluation Metrics}

Based on the evaluation protocols in \cite{alina2020open}, \cite{zhang2019vrd}, \cite{lin2020gps}, we choose the following metrics for the Open Images V6 benchmark: the mean Recall$@50$ ($mR@50$), the regular Recall$@50$ ($R@50$), the weighted mean AP of relationships ($wmAP_{rel}$) and the weighted mean AP of phrases ($wmAP_{phr}$). Like \cite{alina2020open}, \cite{zhang2019vrd}, \cite{lin2020gps}, the weight metric score is defined as: $score_{wtd}=0.2\times R@50 + 0.4\times wmAP_{rel} + 0.4\times wmAP_{phr}$.

\subsubsection{Implementation Details}

As in the case of the Visual Genome experiment, we employ ResNeXt-101-FPN  \cite{he2016deep} as the backbone and for the object detector we choose Faster RCNN \cite{ren2015faster}. Moreover, we freeze the parameters of the above models and apply the same bi-level data resampling strategy\cite{li2021bipartite} as in the previous experiment. The batch size $bs$ is set to 12. Finally, we employ a two-layer MLP to construct the associated log scoring function and utilize an Adam optimizer with the learning rate of $0.0001$.

\subsubsection{Comparisons with State-of-the-Art Methods}

In this experiment, for a fair comparison, several previous methods are re-implemented using the authors's latest code. This is indicated by the $\dagger$ symbol. The results are presented in Table 5. It can be seen that the proposed CSL method achieves  the state-of-the-art performance in all evaluation metrics on the Open Images V6 benchmark. Besides the regular $R@50$ metric, it outperforms the previous methods by a large margin, especially in the more informative $mR@50$ metric, which further verifies the effectiveness of the proposed method.

\subsection{Visualization Results}

In this section, we present typical examples the qualitative results obtained by our proposed CSL method, as well as of the corresponding CSL+BA algorithm, in Fig.7. Specifically, compared with the traditional SGG models, the proposed CSL method is capable of detecting the informative $tail$/$body$ predicate categories rather than the common $head$ predicate categories. For instance, in the top left image, the proposed CSL method detects  informative triplets like $<bird\; standing\; on\; fence>$ and $<cat\; watching\; bird>$. Besides, it can also detect the spatial informative predicates like $in\; front\; of$ or $between$. Moreover, with the balance adjustment strategy, the resulting CSL+BA algorithm further improves its capability in detecting the more informative $tail$ and $body$ predicate categories are. For example,  in the top right image, the resulting CSL+BA method is able to detect more meaningful triplets like $for$ or $growing\; on$, and new additional triplet $<train\; along\; sidewalk>$. As demonstrated in Fig.7, the proposed methods provide much more meaningful structural information, which is expected to benefit downstream tasks, like image captioning or visual question answering.  

\section{Conclusion}

In this paper, we propose a novel constrained structure learning method for the SGG task, in which an explicit constrained variational inference objective is applied in the proposed MFVB framework. Unlike the previous SGG models, the proposed method formulates the SGG task as a more general constrained optimization problem, and investigates an alternative inference technique other than the ubiquitous message passing strategy. Specifically, a generic entropic mirror descent algorithm is applied to accomplish the constrained variational inference step,  while the associated marginals in the proposed MFVB framework are inferred by a specific variable elimination technique. Finally, in extensive experiments on the popular Visual Genome and Open Images V6 benchmarks, we show the proposed generic method outperforms the traditional message passing based SGG models. 

\ifCLASSOPTIONcompsoc
  \section*{Acknowledgments}
\else
  \section*{Acknowledgment}
\fi

This work was supported in part by the U.K. Defence Science and Technology Laboratory, and in part by the Engineering and Physical Research Council (collaboration between U.S. DOD, U.K. MOD, and U.K. EPSRC through the Multidisciplinary University Research Initiative) under Grant EP/R018456/1.

\ifCLASSOPTIONcaptionsoff
  \newpage
\fi

\bibliographystyle{IEEEtran}
\bibliography{IEEEabrv,Semantic}

\end{document}